\documentclass[10pt,twocolumn,letterpaper]{article}

\usepackage[pagenumbers]{cvpr} 

\usepackage{graphicx}
\usepackage{amsmath}
\usepackage{amssymb}
\usepackage{booktabs}
\usepackage{comment}
\usepackage{caption}
\usepackage{stfloats}
\usepackage{times}
\usepackage{epsfig}
\usepackage{ctable}
\usepackage{colortbl}
\usepackage{lipsum}
\usepackage{xcolor}
\usepackage{stfloats}
\usepackage{ulem}

\usepackage{multirow}

\usepackage{subcaption}

\usepackage[accsupp]{axessibility}  

\usepackage[colorlinks=true, citecolor=citecolor, linkcolor=linkcolor, breaklinks=true]{hyperref}
\definecolor{citecolor}{HTML}{0071BC}
\definecolor{linkcolor}{HTML}{ED1C24}

\usepackage[capitalize]{cleveref}
\crefname{section}{Sec.}{Secs.}
\Crefname{section}{Section}{Sections}
\Crefname{table}{Table}{Tables}
\crefname{table}{Tab.}{Tabs.}

\def\cvprPaperID{*****} 
\def\confName{CVPR}
\def\confYear{2023}

\newsavebox\CBox

\newcommand{\OURS}{Emu}
\definecolor{Gray}{gray}{0.92}
\definecolor{darkgreen}{rgb}{0.13, 0.55, 0.13}

\begin{document}


\title{\vspace{-15mm}{\fontsize{13.9}{13.9}\selectfont \OURS: Enhancing Image Generation Models Using Photogenic Needles in a Haystack}\hspace{-0.5cm}\vspace{-5mm}
}

\author{\fontsize{10.6}{12.6}\selectfont
Xiaoliang Dai$^*$, 
Ji Hou$^*$, 
Chih-Yao Ma$^*$, 
Sam Tsai$^*$,
Jialiang Wang$^*$, 
Rui Wang$^*$, 
Peizhao Zhang$^*$,\\
\fontsize{10.6}{12.6}\selectfont
Simon Vandenhende,
Xiaofang Wang, 
Abhimanyu Dubey, 
Matthew Yu, 
Abhishek Kadian, 
Filip Radenovic,  
\\
\fontsize{10.6}{12.6}\selectfont
Dhruv Mahajan,
Kunpeng Li,
Yue Zhao, 
Vladan Petrovic,
Mitesh Kumar Singh,
Simran Motwani,
Yi Wen,
Yiwen Song, 
\\
\fontsize{10.6}{12.6}\selectfont
Roshan Sumbaly$^\dagger$, 
Vignesh Ramanathan$^\dagger$,
Zijian He$^\dagger$,
Peter Vajda$^\dagger$,
Devi Parikh$^\dagger$ \vspace{2mm} \\ 
\fontsize{10.6}{12.6}\selectfont GenAI, Meta \vspace {-1mm}\\
{\tt\footnotesize \{xiaoliangdai, jihou, cyma, sstsai, jialiangw, ruiw, stzpz\}@meta.com} 
}

\twocolumn[{%
	\renewcommand\twocolumn[1][]{#1}%
	\maketitle
	\begin{center}
		\vspace{-1.00cm}
	\includegraphics[width=1.0\linewidth]{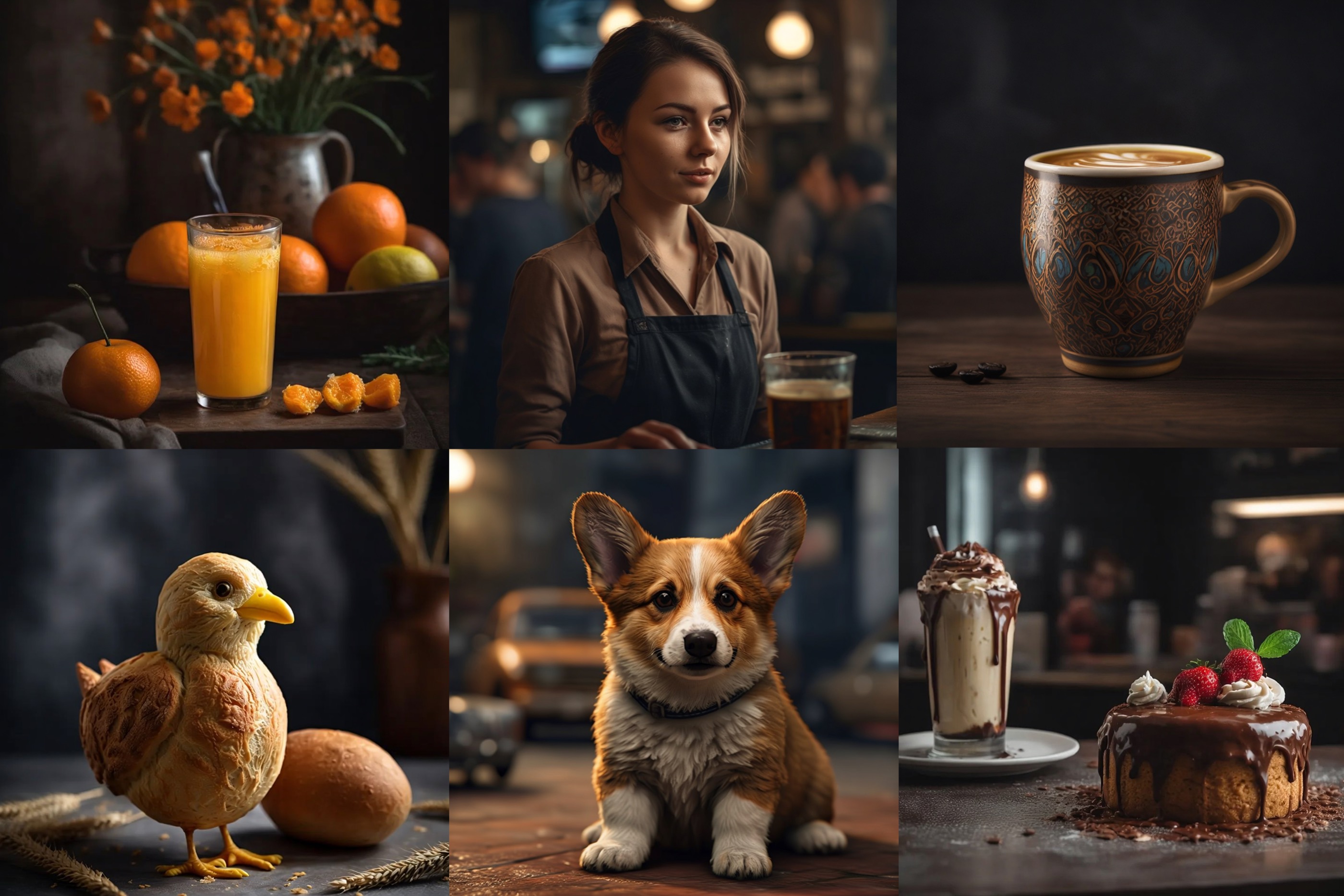}
 \vspace{-7mm}
		\captionof{figure}{With quality-tuning, \OURS{} generates \textit{highly} aesthetic images. Prompts: (top) a glass of orange juice; a woman in an apron works at a local bar; a coffee mug; (bottom) an egg and a bird made of wheat bread; a corgi; a shake is next to a cake. 
  } 
		\label{fig:teaser}
	\end{center}
 \vspace{-2mm}
}]

{\def\iand{\\[5pt]\let\and=\nand}%
 \def\nand{\ifhmode\unskip\nobreak\fi\ $\cdot$ }%
 \let\and=\nand
 \def\at{\\\let\and=\iand}%
 \let\thefootnote\relax\footnote{\kern-\bibindent
 \ignorespaces $^*$ Core contributors: equal contribution, alphabetical order. \\ $^\dagger$ Equal last authors.}\vspace{5dd}}%

\maketitle
\vspace{-0.2cm}
\begin{abstract}{
\vspace{-0.1cm}

Training text-to-image models with web scale image-text pairs enables the generation of a wide range of visual concepts from text. However, these pre-trained models often face challenges when it comes to generating highly aesthetic images. This creates the need for aesthetic alignment post pre-training. In this paper, we propose quality-tuning to effectively guide a pre-trained model to exclusively generate highly visually appealing images, while maintaining generality across visual concepts. Our key insight is that supervised fine-tuning with a set of surprisingly small but extremely visually appealing images can significantly improve the generation quality. We pre-train a latent diffusion model on $1.1$ billion image-text pairs and fine-tune it with only a few thousand carefully selected high-quality images. The resulting model, \OURS, achieves a win rate of $82.9\%$ compared with its pre-trained only counterpart. Compared to the state-of-the-art SDXLv1.0, \OURS{} is preferred $68.4\%$ and $71.3\%$ of the time on visual appeal on the standard PartiPrompts and our Open User Input benchmark based on the real-world usage of text-to-image models. In addition, we show that quality-tuning is a generic approach that is also effective for other architectures, including pixel diffusion and masked generative transformer models.
}
\end{abstract}

\vspace{-5mm}
\section{Introduction}

\label{sec:intro}

\begin{figure*}[h!]
    \centering
    \vspace{-0.5cm}
    \includegraphics[width=1.0\linewidth]{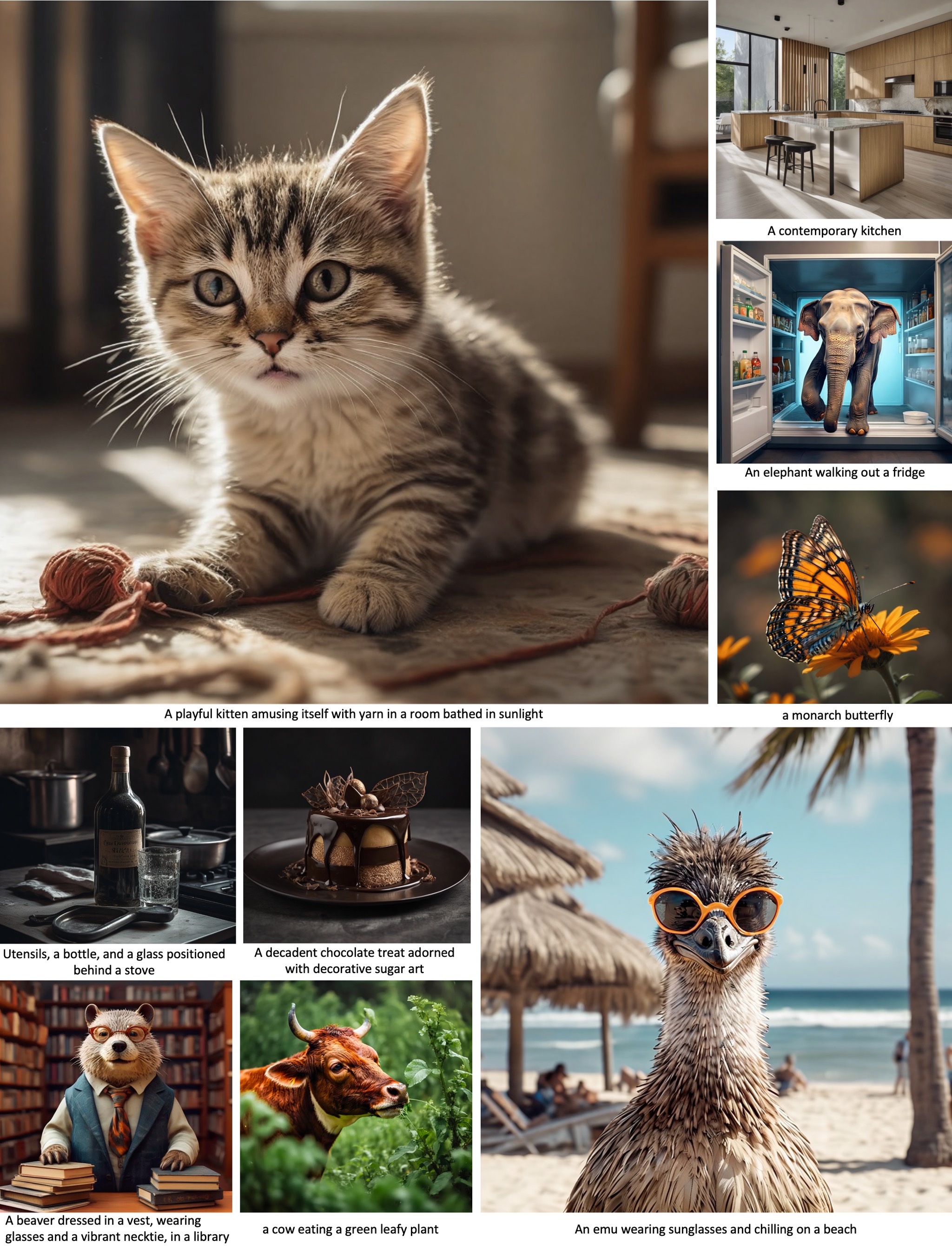}
    \vspace{-0.5cm}
    \caption{\textbf{Selected Examples.} Selected images generated by our quality-tuned model, \OURS.}
    \label{fig:pull_figure}
    \vspace{-1.0cm}
\end{figure*}

Recent advances in generative models have enabled them to generate various high-quality content, such as text~\cite{chatgpt, touvron2023llama}, image~\cite{ramesh2022hierarchical, podell2023sdxl}, music~\cite{huang2023noise2music}, video~\cite{singer2022make}, and even 3D scenes~\cite{poole2022dreamfusion,lin2023magic3d, wang2023prolificdreamer}, which has fundamentally revolutionized generative artificial intelligence (AI). 
In this paper, we present a recipe we have found to be effective for training \textit{highly} aesthetic text-to-image models. It involves two stages: a knowledge learning stage, where the goal is to acquire the ability to generate virtually anything from text, which typically involves pre-training on hundreds of millions of image-text pairs; and a quality learning stage, which is necessary to restrict the output to a high-quality and aesthetically pleasing domain. We refer to the process of fine-tuning for the purpose of improving quality and promoting aesthetic alignment as \textit{quality-tuning} for short.

Our key insight is that to effectively perform quality-tuning, a surprisingly small amount -- a couple of thousand -- exceptionally high-quality images and associated text is enough to make a \textit{significant} impact on the aesthetics of the generated images \textit{without} compromising the generality of the model in terms of visual concepts that can be generated.
Although having more data while maintaining the same level of quality may be helpful, any attempts to prioritize quantity over quality may result in a compromise of the quality of generated images.

This is an interesting finding in the broader landscape of fine-tuning generative models. Quality-tuning for visual generative models can be thought of as analogous to instruction-tuning for large language models (LLMs) in terms of improving generation quality. First, before instruction-tuning, language models are more prone to generating low-quality text, which may be inconsistent in tone, overly verbose or concise, or simply unhelpful or even toxic~\cite{ouyang2022training, bender2021dangers, bommasani2021opportunities, weidinger2021ethical, gehman2020realtoxicityprompts}; ChatGPT-level~\cite{chatgpt} performance is achieved with effective instruction-tuning~\cite{ouyang2022training}. Similarly, we find that quality-tuning significantly improves the generation quality of text-to-image models. Second, the recipe for effectively performing instruction-tuning and quality-tuning is similar: use high-quality data, even if the quantity has to be small to maintain quality. Llama2~\cite{touvron2023llama} has been fine-tuned on $27$K high-quality prompts, which can be considered a very small quantity compared to the billions or trillions of pre-training tokens. 
Similarly, we find that strong text-to-image performance can be achieved by fine-tuning with even less data -- a few thousand carefully selected images. Lastly, the knowledge obtained from pre-training is mostly retained after both instruction-tuning and quality-tuning. Like instruction-tuned LLMs, quality-tuned text-to-image models retain their generality in terms of the visual concepts that can be generated. These post pre-training stages align the knowledge to downstream user value -- improving text quality and following instructions in the case of LLMs, and promoting aesthetic alignment in the case of text-to-image models.

Concretely, we pre-train a latent diffusion model (LDM) on $1.1$ billion image-text pairs and quality-tune the model on a few thousand hand-picked exceptionally high-quality images selected from a large corpus of images.  By its nature, the selection criterion is subjective and culturally dependent. We follow some common principles in photography, including but not limited to composition, lighting, color, effective resolution, focus, and storytelling to guide the selection process. 
With a few optimizations to the latent diffusion architecture, we start with a strong pre-trained model and dramatically improve the visual appeal of our generated images through quality-tuning. In fact, it significantly outperform a state-of-the-art publicly available model SDXLv1.0~\cite{podell2023sdxl} on visual appeal. 
We call our quality-tuned LDM \OURS. We show example generations from \OURS~in Figure~\ref{fig:teaser} and Figure~\ref{fig:pull_figure}. 

Furthermore, we show that quality-tuning is a generic approach that is also effective for pixel diffusion and masked generative transformer models.

Our main contributions are:

\begin{itemize}
    \item We build \OURS, a quality-tuned latent diffusion model that significantly outperforms a publicly available state-of-the-art  model SDXLv1.0 on visual appeal. 
    \item To the best of our knowledge, this is the first work to emphasize the importance of a good fine-tuning recipe for aesthetic alignment of text-to-image models. We provide insights and recommendations for a key ingredient of this recipe -- supervised fine-tuning with a surprisingly small amount of exceptionally high-quality data can have a significant impact on the quality of the generated images. Image quality should always be prioritized over quantity.
    \item We show that quality-tuning is a generic approach that also works well for other popular model architectures besides LDM, including pixel diffusion and masked generative transformer models. 
    
\end{itemize}

\section{Related Work}

\paragraph{Text-to-Image Models.} Generating an image from a textual description has been explored using various approaches. Diffusion-based methods learn a denoising process to gradually generate images from pure Gaussian noise~\cite{ho2020denoising}. The denoising process can occur either in pixel space or latent space, resulting in pixel diffusion models~\cite{ramesh2022hierarchical, saharia2022photorealistic, balaji2022ediffi} or latent diffusion models~\cite{rombach2022high, podell2023sdxl}, which feature higher efficiency by reducing the size of spatial features. Generative transformer methods usually train autoregressive~\cite{ramesh2021zero, yu2022scaling, gafni2022make, ding2021cogview, aghajanyan2022cm3, yu2023scaling} or non-autoregressive (masked)~\cite{chang2023muse} transformers in discrete token space to model the generation process. Generative adversarial network~\cite{sauer2023stylegan, kang2023scaling} also show remarkable capability of generating realistic images. While these models exhibit unprecedented image generation ability, they do not \textit{always} generate \textit{highly} aesthetic images.

\paragraph{Fine-Tuning Text-to-Image Models.}
Given a pre-trained text-to-image model, different methods have been developed to enable specific tasks. A number of techniques have been developed to personalize or adapt text-to-image models to a new subject or style~\cite{ruiz2022dreambooth, gal2022image, hu2021lora}. ControlNet~\cite{zhang2023adding} provides additional control to the generation process by additionally conditioning on pose, sketch, edges, depth, etc. 
InstructPix2Pix~\cite{brooks2023instructpix2pix} makes text-to-image models follow editing instructions by fine-tuning them on a set of generated image editing examples. To the best of our knowledge, this is the first work highlighting fine-tuning for generically promoting aesthetic alignment for a wide range of visual domains.

\paragraph{Fine-Tuning Language Models.} Fine-tuning has become a critical step in building high-quality LLMs~\cite{touvron2023llama, ouyang2022training, gpt4}. It generically improves output quality while enabling instruction-following capabilities. Effective fine-tuning of LLMs can be achieved with a relatively small but high-quality fine-tuning dataset, \textit{e.g.,} using $27$K prompts in~\cite{touvron2023llama}. 
In this work, we show that effective fine-tuning of text-to-image models can be also achieved with a \textit{small} but \textit{high-quality} fine-tuning dataset. This finding shows an interesting connection between fine-tuning vision and language models in generative AI.

\section{Approach}
\label{sec:approach}

As discussed earlier, our approach involves a knowledge learning stage followed by a quality-tuning stage. This may seem like a well-known recipe when mapped to pre-training and fine-tuning. That said, the key insights here are:  (i) the fine-tuning dataset can be surprisingly small, on the order of a couple of thousand images, (ii) the quality of the dataset needs to be very high, making it difficult to fully automate data curation, requiring manual annotation, and (iii) even with a small fine-tuning dataset, quality-tuning not only significantly improves the aesthetics of the generated images, but does so without sacrificing generality as measured by faithfulness to the input prompt. Note that the stronger the base pre-trained model, the higher the quality of the generated images after quality-tuning. To this end, we made several modifications to the latent diffusion architecture~\cite{rombach2022high} to facilitate high-quality generation. That said, quality-tuning is general enough and can be applied to a variety of architectures.
 
In this section, we first introduce the latent diffusion architecture we use. Then, we discuss the pre-training stage, followed by the protocol for collecting the high-quality fine-tuning dataset, and finally the quality-tuning stage. Later in Section~\ref{sec:results}, we demonstrate that quality-tuning is not limited to latent diffusion models but also improve other models such as pixel diffusion~\cite{saharia2022photorealistic} and masked generative transformer~\cite{chang2023muse} models.

\begin{figure}[t!]
    \centering
    \includegraphics[width=0.97\linewidth]{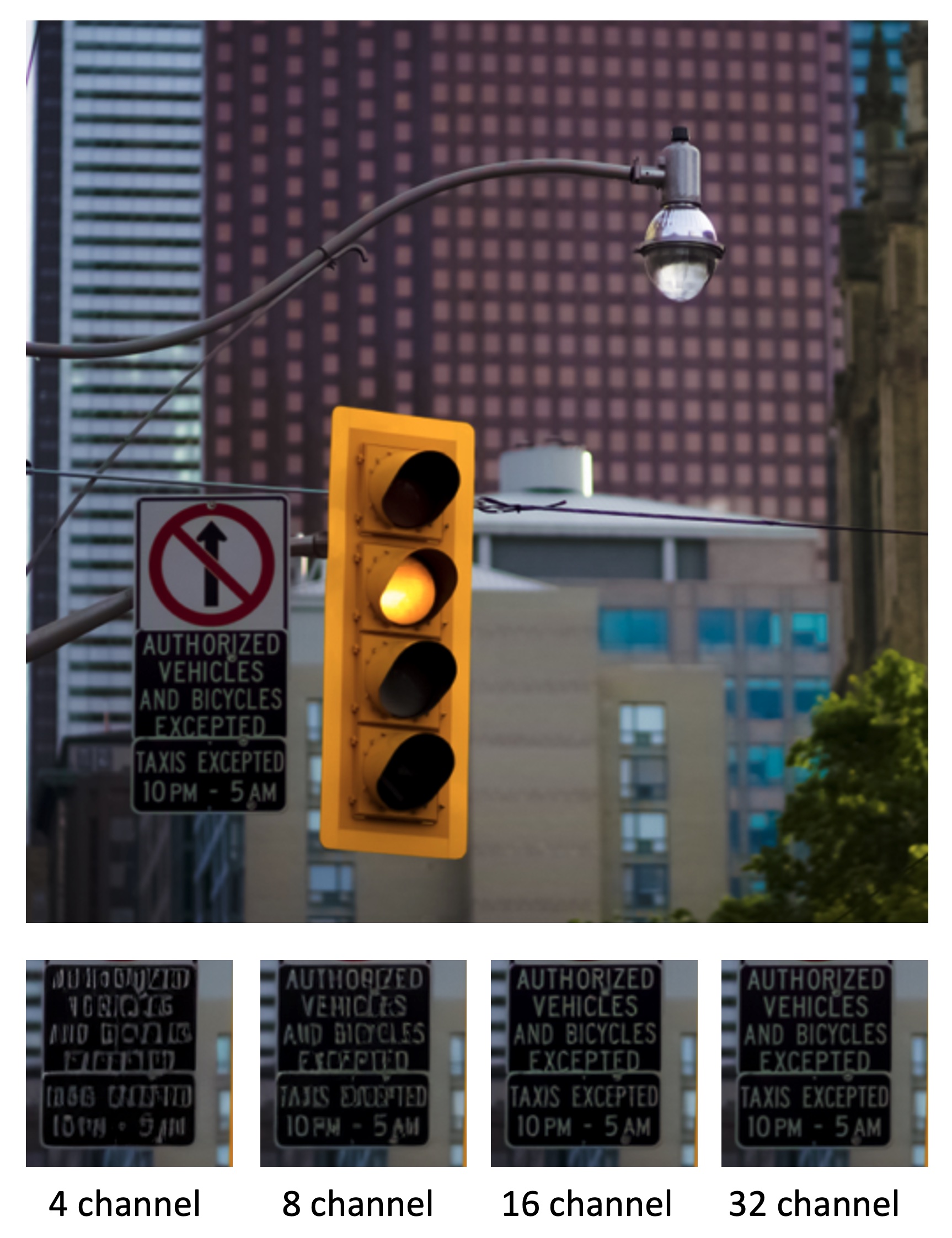}
    \caption{\textbf{Autoencoder.} The visual quality of the reconstructed images for autoencoders with different channel sizes. While keeping all other architecture layers the same, we only change the latent channel size. We show that the original 4-channel autoencoder design~\cite{rombach2022high} is unable to reconstruct fine details. Increasing channel size leads to much better reconstructions. We choose to use a 16-channel autoencoder in our latent diffusion model.}
    \label{fig:ae}
\end{figure}

\subsection{Latent Diffusion Architecture}
We design a latent diffusion model that outputs $1024 \times 1024$ resolution images. Following standard latent diffusion architecture design, our model has an autoencoder (AE) to encode an image to latent embeddings and a U-Net to learn the denoising process.

We find that the commonly used 4-channel autoencoder (AE-4) architecture often results in a loss of details in the reconstructed images due to its high compression rate. 
The issue is especially noticeable in small objects. Intuitively, it compresses the image resolution by $64\times$ with three $2\times2$ downsampling blocks but increases the channel size only from 3 (RGB) to 4 (latent channels). We find that increasing the channel size to 16 significantly improves reconstruction quality (see Table~\ref{tab:ae_c_size}). To further improve the reconstruction performance, we use an adversarial loss and apply a non-learnable pre-processing step to RGB images using a \textit{Fourier Feature Transform} 
to lift the input channel dimension from 3 (RGB) to a higher dimension to better capture fine structures.
See Figure~\ref{fig:ae} for qualitative results of autoencoders of different channel size.

\begin{table}[h!]
\centering
\small
\begin{tabular}{lccccc}
model & channel & SSIM & PSNR & FID \\
\toprule
& 4 & 0.80 & 28.64 & 0.35\\
AE & 8 & 0.86 & 30.95 & 0.19 \\
& 16 & 0.92 & 34.00 & 0.06 \\
\hline
Fourier-AE & 16 & 0.93 & 34.19 & 0.04 \\
\end{tabular}
\caption{While keeping all other architecture design choices fixed, we first only change the latent channel size and report their reconstruction metrics on ImageNet~\cite{russakovsky2015imagenet}. We see that AE-16 significantly improves over AE-4 on all reconstruction metrics. Adding a Fourier Feature Transform and an adversarial loss further improves the reconstruction performance.}
\label{tab:ae_c_size} 
\end{table}

We use a large U-Net with 2.8B trainable parameters. We increase the channel size and number of stacked residual blocks in each stage for larger model capacity. We use text embeddings from both CLIP ViT-L~\cite{radford2021learning} and T5-XXL~\cite{raffel2020exploring} as the text conditions.

\subsection{Pre-training}
We curate a large internal pre-training dataset consisting of $1.1$ billion images to train our model. 
The model is trained with progressively increasing resolutions, similar to~\cite{podell2023sdxl}.
This progressive approach allows the model to efficiently learn high-level semantics at lower resolutions and improve finer details at the highest resolutions 
We also use a noise-offset~\cite{noiseoffset} of 0.02 in the final stage of pre-training. This facilitates high-contrast generation, which contributes to the aesthetics of the generated images. 

\begin{figure}[t!]
    \centering
    \includegraphics[width=1.0\linewidth]{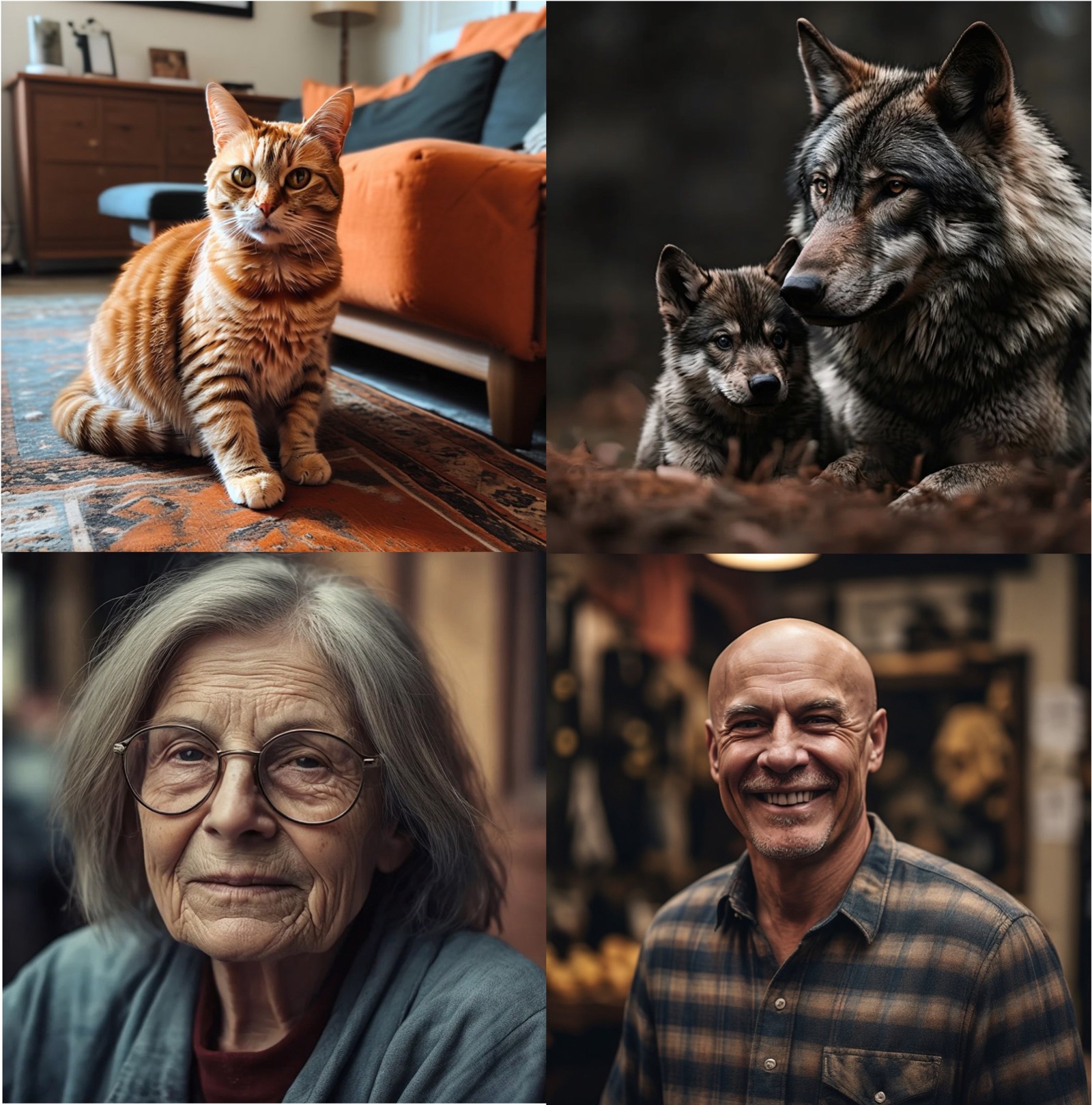}
    \caption{\textbf{Visual Appealing Data}. Examples of visually appealing data that can meet our human filtering criterion. 
    }
    \label{fig:example_finetuning_data}
\end{figure}

\subsection{High-Quality Alignment Data}
As discussed before, in order to align the model towards highly aesthetic generations -- quality matters significantly more than quantity in the fine-tuning dataset (see Section~\ref{sec:results:ablation} for an ablation study on quality vs quantity). As also discussed, the notion of aesthetics is highly subjective. Here we discuss in detail what aesthetics we chose and how we curated our fine-tuning dataset by combining both automated filtering and manual filtering. The general quality-tuning strategy will likely apply to other aesthetics as well.

\noindent \textbf{Automatic Filtering.} Starting from an initial pool of billions of images, we first use a series of automatic filters to reduce the pool to a few hundreds of millions. These filters include but are not limited to offensive content removal, aesthetic score filter, optical character recognition (OCR) word count filter to eliminate images with too much overlaying text on them, and CLIP score filter to eliminate samples with poor image-text alignment, which are standard pre-filtering steps for sourcing large datasets. We then perform additional automated filtering via image size and aspect ratio. Lastly, to balance images from various domains and categories, we leverage visual concept classification~\cite{yalniz2019billion} to source images from specific domains (\textit{e.g.,} portrait, food, animal, landscape, car, etc). Finally, with additional quality filtering based on proprietary signals (\textit{e.g.}, number of likes), we can further reduce the data to $200$K. 

\noindent \textbf{Human Filtering.} 
Next, we perform a two-stage human filtering process to only retain highly aesthetic images. In the first stage, we train \textit{generalist} annotators to downselect the image pool to 20K images. Our primary goal during this stage is to optimize recall, ensuring the exclusion of medium and low quality that may have passed through the automatic filtering.
In the second stage, we engage \textit{specialist} annotators who have a good understanding of a set of photography principles. Their task is to filter and select images of the highest aesthetic quality (see Figure~\ref{fig:example_finetuning_data} for examples). During this stage, we focus on optimizing precision, meaning we aim to select only the very best images.
A brief annotation guideline for photorealistic images is as follows. Our hypothesis is that following basic principles of high quality photography leads to generically more aesthetic images across a variety of styles, which is validated via human evaluation.
\begin{enumerate}
    \item \textbf{Composition.} The image should adhere to certain principles of professional photography composition, including the ``Rule Of Thirds'', ``Depth and Layering'', and more. Negative examples may include imbalance in visual weight, such as when all focal subjects are concentrated on one side of the frame, subjects captured from less flattering angles, or instances where the primary subject is obscured, or surrounding unimportant objects are distracting from the subject. 

    \item \textbf{Lighting.} We are looking for dynamic lighting with balanced exposure that enhances the image, for example, lighting that originates from an angle, casting highlights on select areas of the background and subject(s). We try to avoid artificial or lackluster lighting, as well as excessively dim or overexposed light. 

    \item \textbf{Color and Contrast.} We prefer images with vibrant colors and strong color contrast. We avoid monochromatic images or those where a single color dominates the entire frame.
    \item \textbf{Subject and Background.} The image should have a sense of depth between the foreground and background elements. The background should be uncluttered but not overly simplistic or dull. The focused subjects must be intentionally placed within the frame, ensuring that all critical details are clearly visible without compromise. For instance, in a portrait, the primary subject of image should not extend beyond the frame or be obstructed. Furthermore, the level of detail on the foreground subject is extremely important. 

    \item \textbf{Additional Subjective Assessments.} Furthermore, we request annotators to provide their subjective assessments to ensure that only images of \textit{exceptionally} aesthetic quality are retained by answering a couple of questions, such as:
    (i) Does this image convey a compelling story?
    (ii) Could it have been captured significantly better?
    (iii) Is this among the best photos you've ever seen for this particular content?
\end{enumerate}
 
Through this filtering process, we retained a total of 2000 exceptionally high-quality images. Subsequently, we composed ground-truth captions for each of them. Note that some of these handpicked images are below our target resolution of $1024 \times 1024$. 
We trained a pixel diffusion upsampler inspired by the architecture proposed in~\cite{saharia2022photorealistic} to upsample these images when necessary. 

\subsection{Quality-Tuning}
We can think of the visually stunning images (like the 2000 images we collected) as a subset of all images that share some common statistics.
Our hypothesis is that a strongly pre-trained model is already capable of generating highly aesthetic images, but the generation process is not properly guided towards always producing images with these statistics. Quality-tuning effectively restricts outputs to a high-quality subset. 

We fine-tune the pre-trained model with a small batch size of $64$. 
We use a noise-offset of $0.1$ at this stage. Note that early stopping is important here as fine-tuning on a small dataset for too long will result in significant overfitting and degradation in generality of visual concepts. We fine-tune for no more than $15$K iterations despite the loss still decreasing. This total iteration number is determined empirically.

\section{Experiments}
\label{sec:results}

We compare our quality-tuned model to our pre-trained model to demonstrate the effectiveness of quality-tuning. To place the visual appeal of our generated images in context with a current state-of-the-art model, we compare our model to SDXLv1.0~\cite{podell2023sdxl}. Due to lack of access to training data of SDXL and their underlying model, we leveraged their corresponding APIs for our comparison. Note that unlike SDXL, we use a single stage architecture, and do not use a subsequent refinement stage. As stated earlier, we also show that quality-tuning is not specific to LDMs, and can be applied to other architectures -- pixel diffusion and masked generative transformer models.

\subsection{Evaluation Setting}
\begin{figure}[t!]
    \centering
    \includegraphics[width=0.95\linewidth]{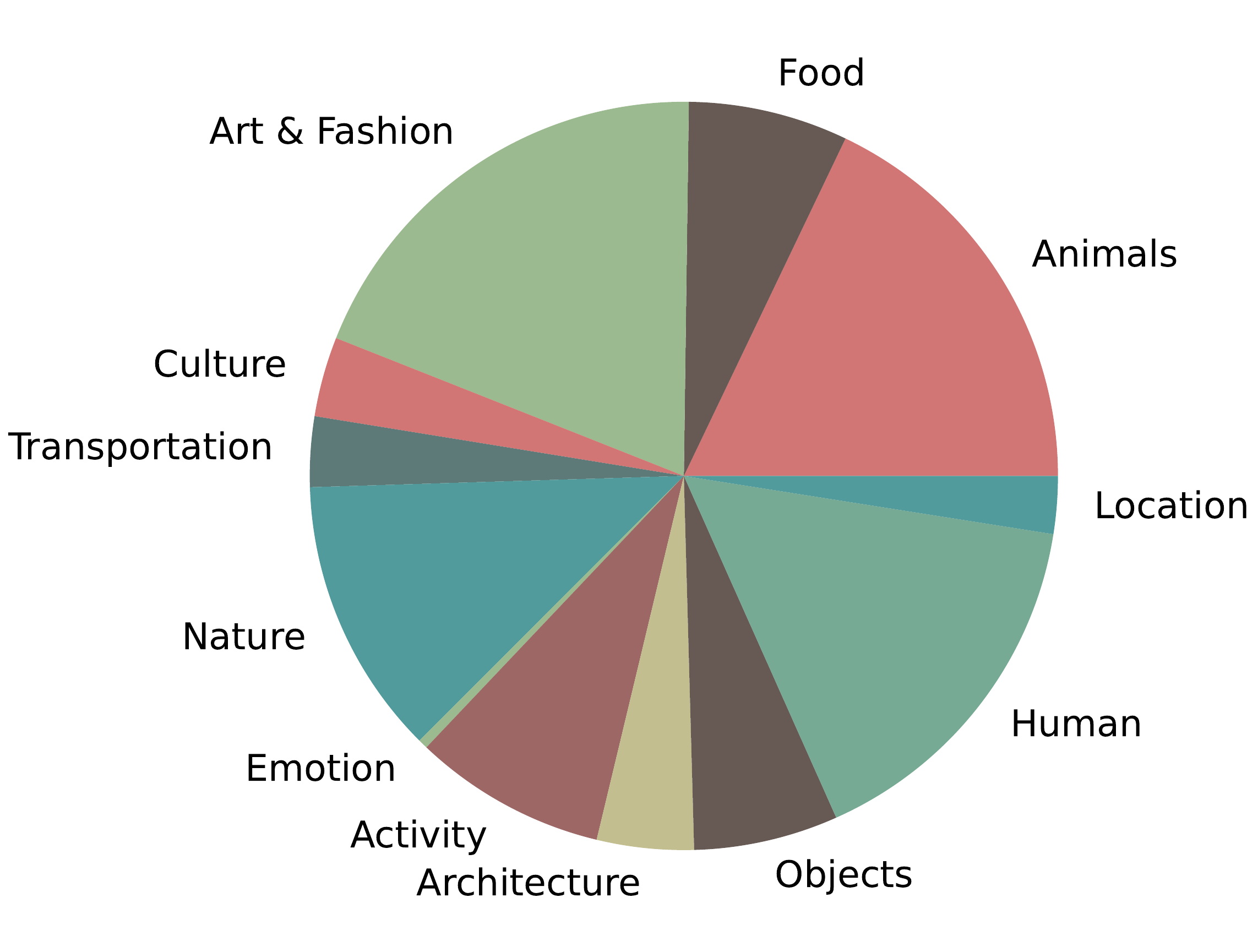}
    \caption{\textbf{Prompt distributions}. The distribution of different concepts in our Open User Input prompts. We cover a comprehensive list of common concepts people typically use to generate images. 
    }
    \label{fig:distribution}
\end{figure}

\paragraph{Prompts.}
We evaluate on two large sets of prompts: 1600 PartiPrompts~\cite{yu2022scaling} which is commonly used for text-to-image generation benchmarking, and our 2100 Open User Input (OUI) Prompts. 
The OUI prompt set is based on real-world user prompts. It captures prompts that are popular with text-to-image models, reflecting visual concepts that are relevant to real-world use cases (Figure~\ref{fig:distribution}), paraphrased by LLMs to be closer to how regular users might input prompts (as opposed to being highly prompt engineered). The overall motivation was to capture the creativity and intricacies of popular prompts for text-to-image models so we are pushing the capabilities of models, while also being grounded in likely real world use cases. 

\paragraph{Metrics.}
We use two separate evaluation metrics: visual appeal and text faithfulness. 
Visual appeal refers to the overall aesthetic quality of a generated image. It combines various visual elements such as color, shape, texture, and composition that creates a pleasing and engaging look. The concept of visual appeal is subjective and varies from person to person, as what may be aesthetically pleasing to one person may not be to another.  Therefore, we ask five annotators to rate each sample. Concretely, we show annotators two images A and B, side-by-side, each generated by a different model using the same caption. The text captions are not displayed. Annotators choose which image is more visually appealing by selecting ``A'', ``B'' or ``Tie''.

Text faithfulness refers to the degree of similarity between a generated image and a text caption. In this task, we again display two generated images A and B, side-by-side, but with the caption alongside the images. The annotators are asked to ignore the visual appeal of the images and choose which ones best describe the caption with choices ``A '', ``B'', ``Both'', and ``Neither'', where ``Both'' and ``Neither'' are considered as ``Tie''.
In this task, we have three annotators to annotate each sample pair. 

We do not report ``standard'' metrics such as FID scores. As argued in many recent papers (\textit{e.g.,}~\cite{podell2023sdxl, kirstain2023pick}), FID scores do not correlate well with human assessment of the performance of generative models. 

\begin{figure}[t!]
    \centering
    \begin{subfigure}[b]{0.48\textwidth}
    \includegraphics[width=1\linewidth]{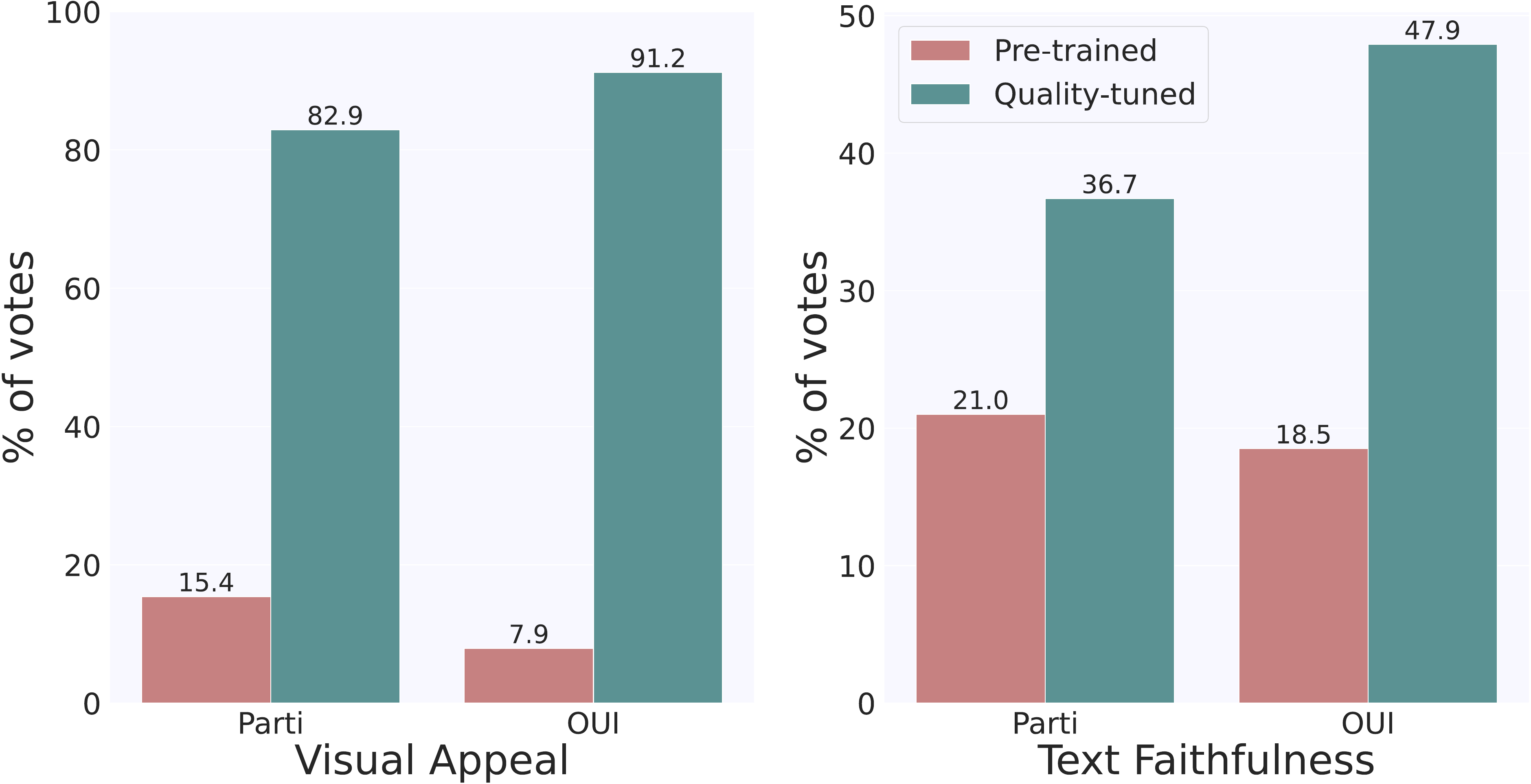}
    \caption{All Prompts}
    \vspace{0.35cm}
    \end{subfigure}
    \begin{subfigure}[b]{0.48\textwidth}
    \includegraphics[width=1\linewidth]{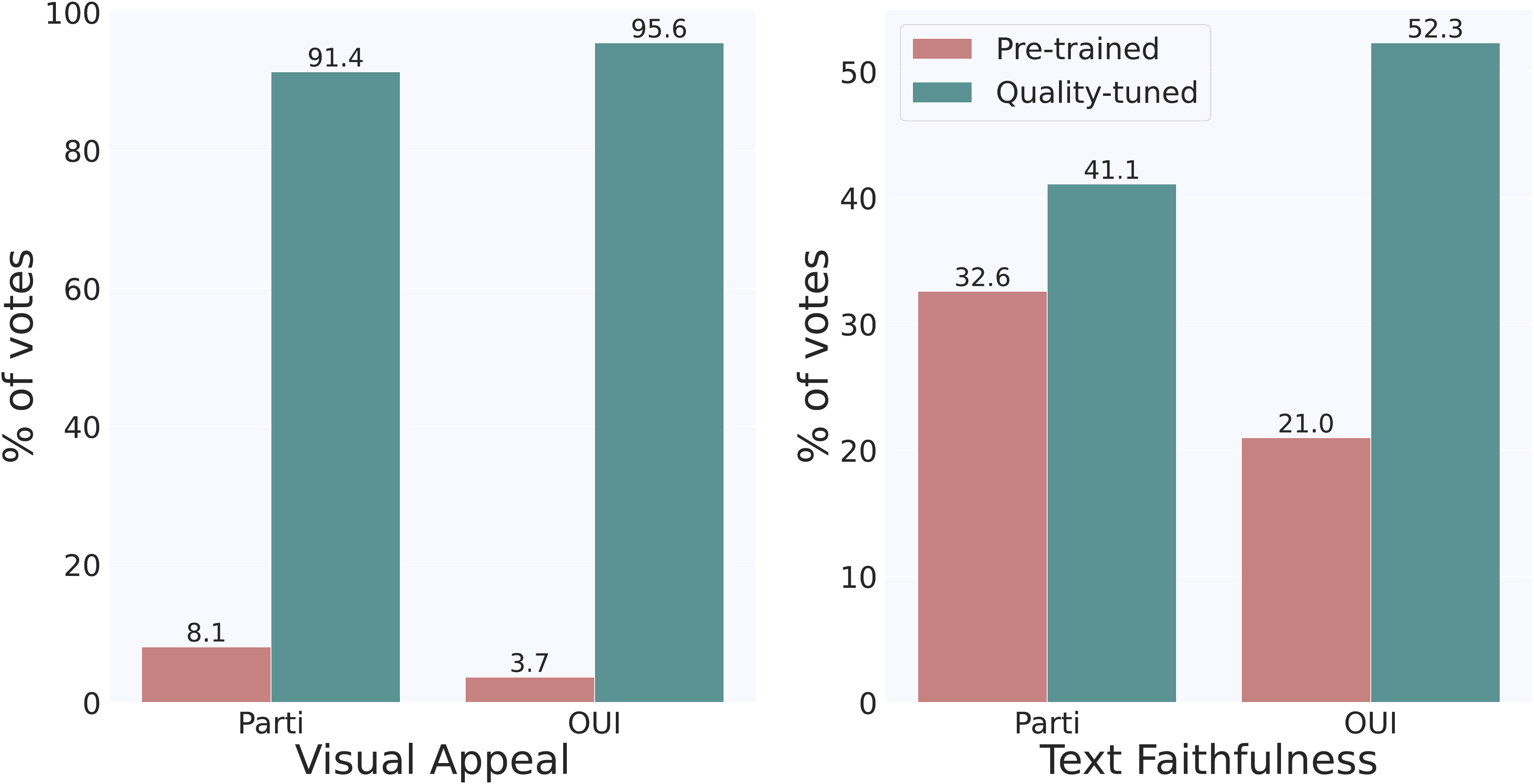}
    \caption{Stylized Prompts}
    \end{subfigure}
    \caption{\textbf{Quality-Tuning vs Pre-training}. Human evaluation on both the PartiPrompts and Open User Input prompts shows that our quality-tuned model, \OURS, significantly outperforms the pre-trained 
    model on visual appeal, without loss of generality of visual concepts or styles that can be generated. 
    }
    \label{fig:pretrain_vs_finetune}
\end{figure}

\begin{figure}[h!]
    \centering
    \vspace{-0.2cm}
    \includegraphics[width=1.0\linewidth]{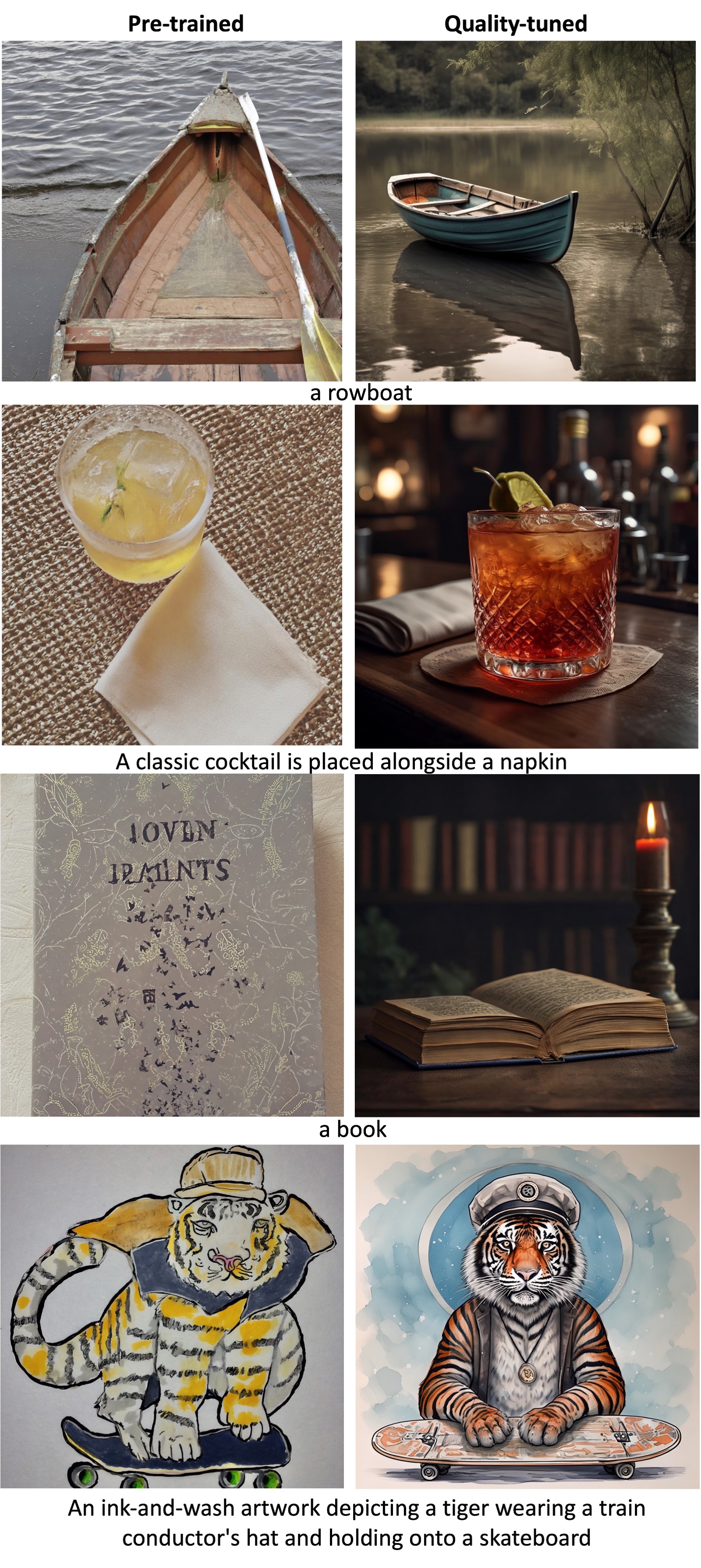}
    \caption{\textbf{Qualitative Comparison}. a comparison of images generated by the pre-trained and quality-tuned model.
    } 
    \label{fig:qualitative_results}
\end{figure}

\subsection{Results}
\begin{figure*}[h!]
    \centering
    \vspace{-1.2cm}
    \includegraphics[width=0.97\linewidth]{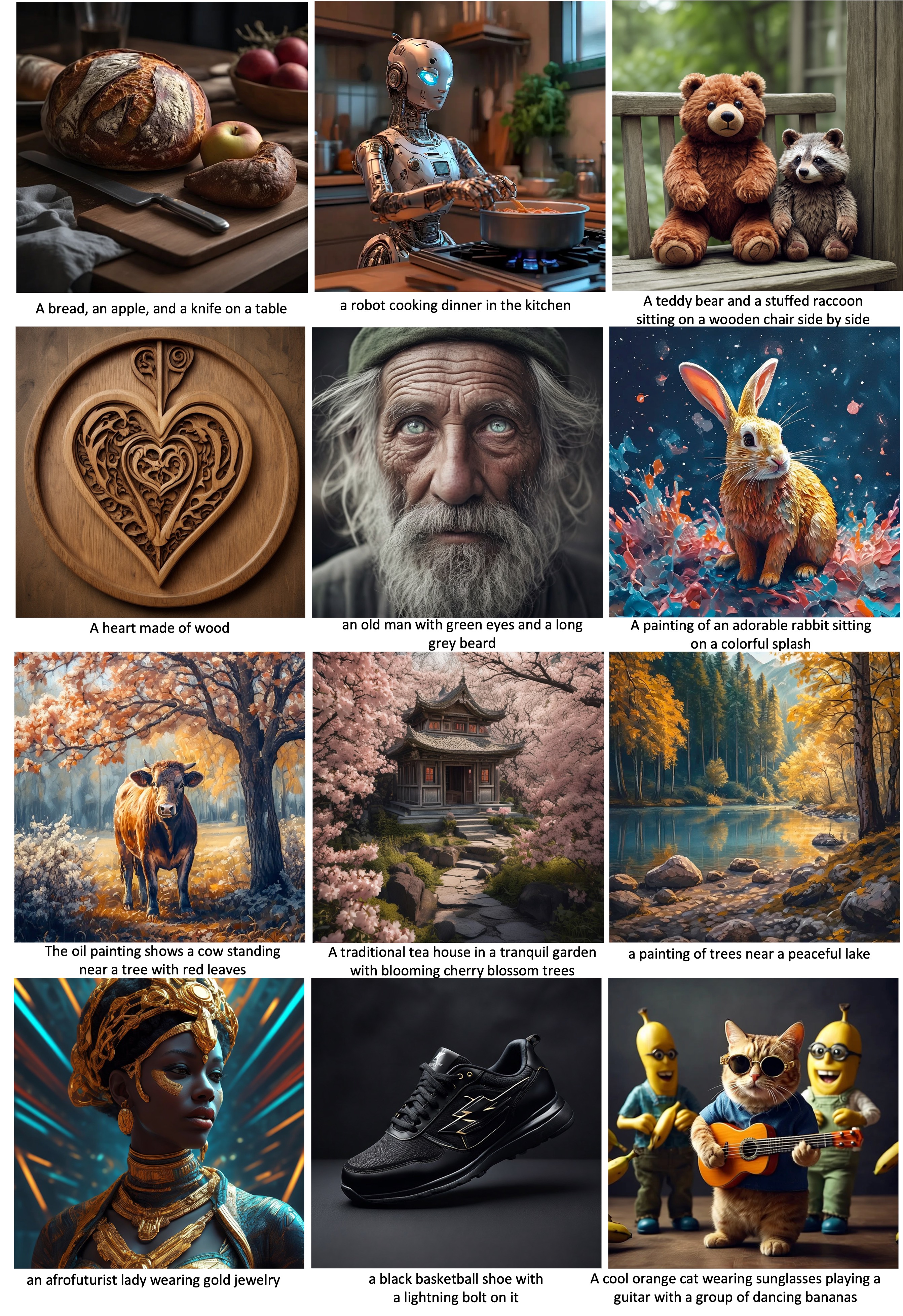}
    \vspace{-0.5cm}
    \caption{\textbf{Generated Examples.} Images generated by our quality-tuned model, \OURS.} 
    \label{fig:more_examples}
    \vspace{-1.2cm}
\end{figure*}

\begin{figure*}[h!]
    \centering
    \vspace{-1.2cm}
    \includegraphics[width=0.97\linewidth]{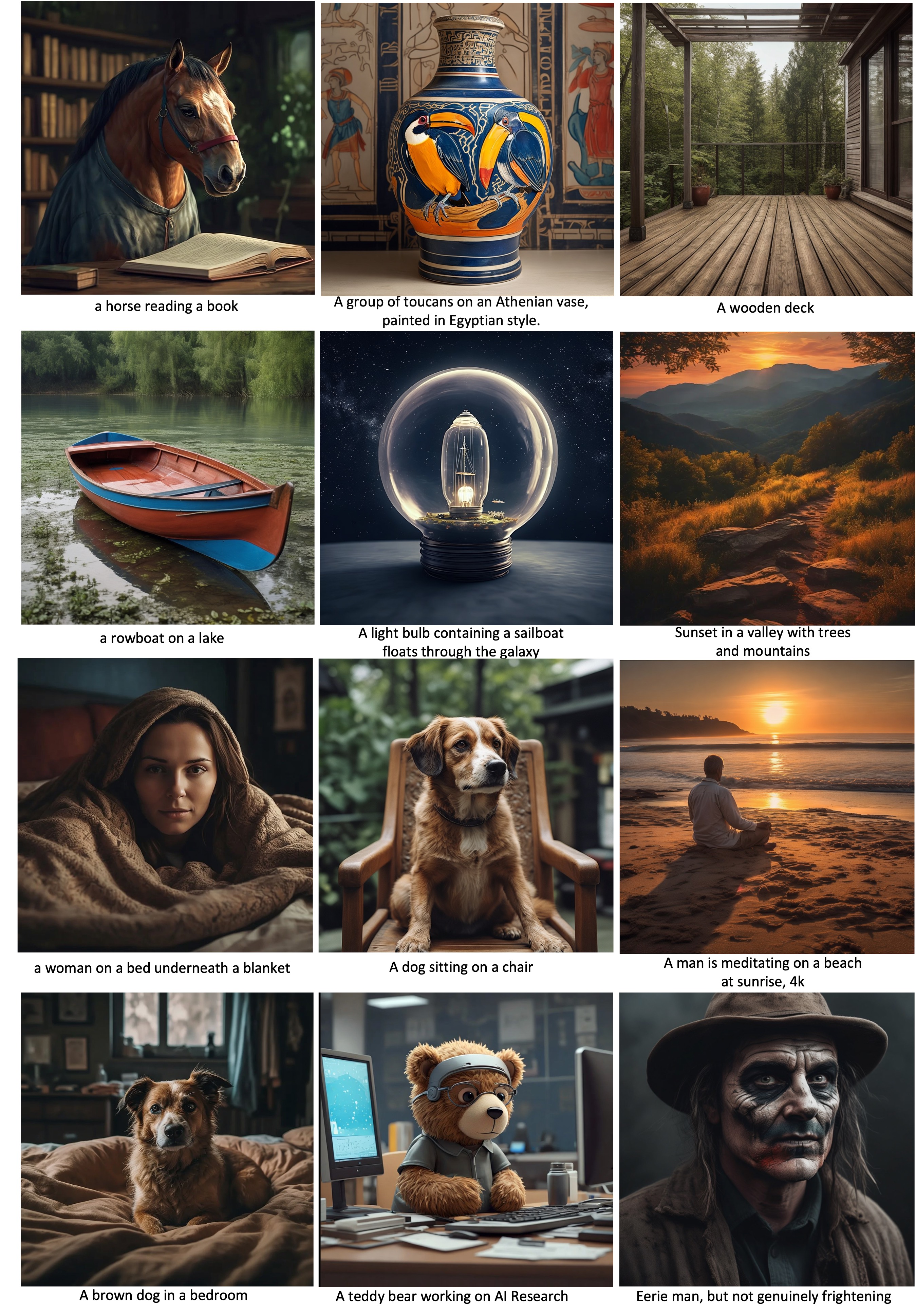}
    \vspace{-0.5cm}
    \caption{\textbf{More Examples.} Images generated by our quality-tuned model, \OURS.} 
    \label{fig:more_examples_2}
    \vspace{-1.2cm}
\end{figure*}

\paragraph{Effectiveness of Quality-Tuning.} First, we compare our quality-tuned model, \OURS, with the pre-trained model. See Figure~\ref{fig:qualitative_results} for random (not cherry-picked) qualitative examples before and after quality-tuning. Note the highly aesthetic non-photorealistic image as well, validating our hypothesis that following certain photography principles in curating the quality-tuning dataset leads to improved aesthetics for a broad set of styles.
We show more examples of generated images using \OURS{} in Figure~\ref{fig:more_examples} and Figure~\ref{fig:more_examples_2}.

Quantitatively, as shown in Figure~\ref{fig:pretrain_vs_finetune} (top), 
after quality-tuning, \OURS{} is preferred in both visual appeal and text faithfulness by a significant margin. 
Specifically, \OURS{} is preferred $82.9\%$ and $91.2\%$ of the time for visual appeal, and $36.7\%$ and $47.9\%$ of the time for text faithfulness on PartiPrompts and OUI Prompts, respectively. In contrast, the pre-trained model is preferred only $15.4\%$ and $7.9\%$ of the time for visual appeal, and $21.0\%$ and $18.5\%$ of the time for text faithfulness on PartiPrompts and OUI Prompts. The remaining cases result in ties. 
From the two large sets of evaluation data that covers various domains and categories, we did not observe degradation of generality across visual concepts. In fact, as seen, text-faithfulness also improved. We hypothesize this is because the captions of the 2000 quality-tuning images were manually written, while the captions in the pre-training dataset tend to be noisy. Finally, we analyze the results on non-photorealistic stylized prompts (\textit{e.g.,} sketches, cartoons, etc.). We find that the improvements broadly apply to a variety of styles, see Figure~\ref{fig:pretrain_vs_finetune} (bottom).

\paragraph{Visual Appeal in the Context of SoTA.}
To place the visual appeal of our generated images in the context of current state-of-the-art, we compare \OURS{} with SDXLv1.0~\cite{podell2023sdxl}. As shown in Table~\ref{tab:sdxl_visual}, our model is preferred over  SDXLv1.0 in visual appeal by a significant margin -- including on stylized (non-photorealistic) prompts.

\begin{table}[h!]
\small
  \centering
  \addtolength{\tabcolsep}{-1pt}
  \begin{tabular}{lccc}
    Eval data &  win  ($\%$) & tie ($\%$) & lose ($\%$) \\
    \toprule
    Parti (All) & 68.4 & 2.1 & 29.5 \\
    OUI  (All) & 71.3 & 1.2 & 27.5\\ 
    Parti (Stylized) & 81.7 & 1.9 & 16.3\\
    OUI  (Stylized) & 75.5 & 1.4 & 23.1\\ 
  \end{tabular}
  \caption{\textbf{\OURS{} vs SDXL on Visual Appeal}. Our model is preferred over SDXL by a large margin, including on stylized, non-photorealistic prompts.
  }
\label{tab:sdxl_visual}
\end{table}

\begin{figure}[h!]
    \centering
    \includegraphics[width=1\linewidth]{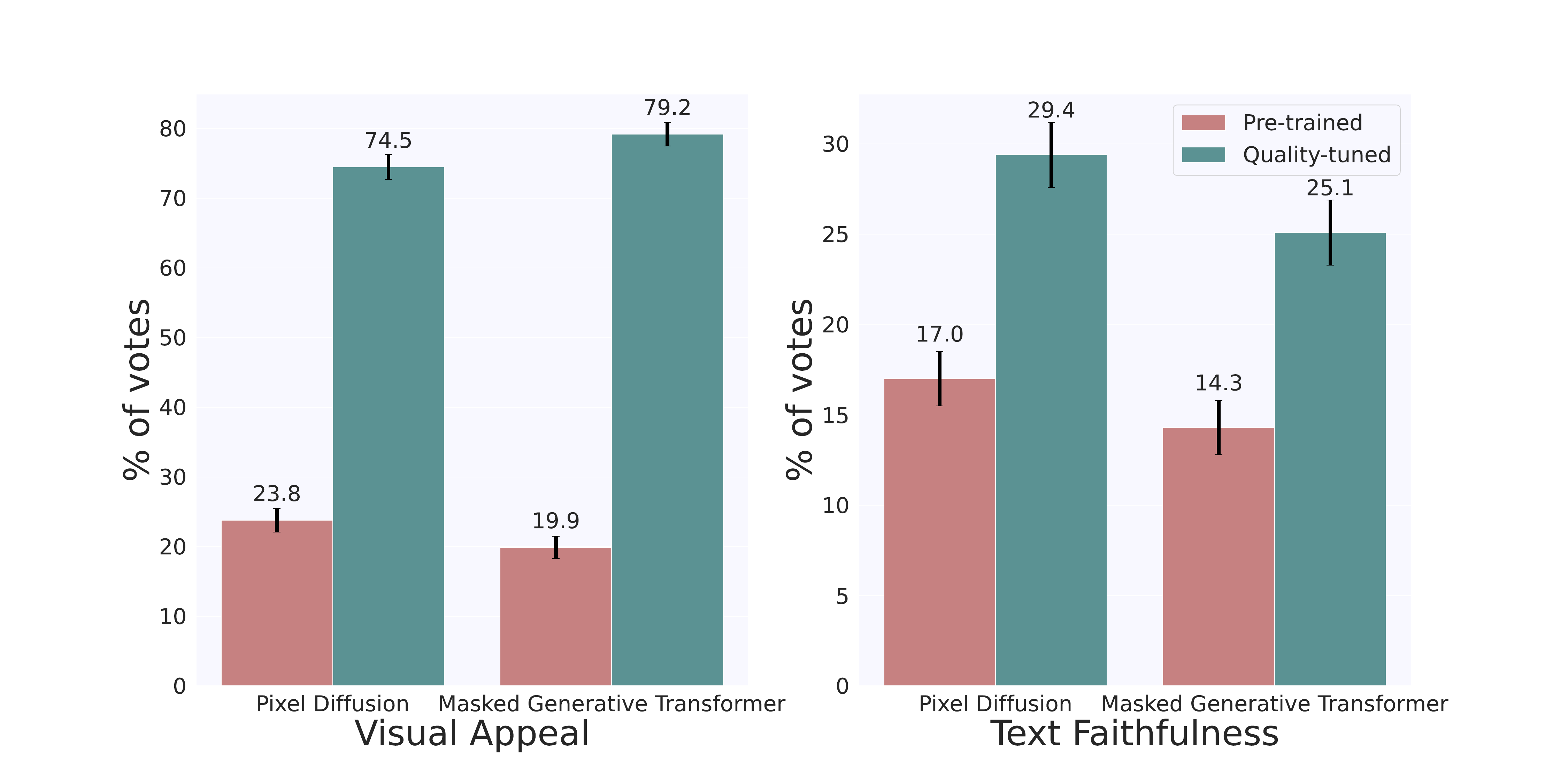}
    \caption{
        \textbf{Quality-Tuning vs Pre-training on Pixel Diffusion and Masked Generative Transformer}. We adapt our quality-tuning to other popular text-to-image model architectures. Our results indicate that the success of quality-tuning can be transferred to other architectures, beyond latent diffusion models.
    }
    \vspace{-0.3cm}
    \label{fig:other_archs}
\end{figure}

\paragraph{Quality-Tuning Other Architectures.} 
Next, we show our quality-tuning can also improve other popular architectures, such as pixel diffusion and masked generative transformer models. 
Specifically, we re-implement and re-train from scratch a pixel diffusion model, similar to Imagen~\cite{saharia2022photorealistic} architecture, and a masked generative transformer model, similar to Muse~\cite{chang2023muse} architecture, and then quality-tune them on 2000 images. 
We evaluate both quality-tuned models on $1/3$ randomly sampled PartiPrompts. 
As shown in Figure~\ref{fig:other_archs}, both architectures show significant improvement after quality-tuning on both visual appeal and text faithfulness metrics.

\subsection{Ablation Studies}
\label{sec:results:ablation}

We do ablation studies on the fine-tuning dataset with a focus on visual appeal. We first investigate the impact of the dataset size. We report results of quality-tuning on randomly sampled subsets of sizes 100, 1000 and 2000 in Table~\ref{tab:dataset_size_visual}.
With even just 100 fine-tuning images, the model can already be guided towards generating visually-pleasing images, jumping from a win rate of $24.8\%$ to $60\%$ compared with SDXL. 

\begin{table}[h!]
\small
  \centering
  \begin{tabular}{l|ccc}
   fine-tune data & win ($\%$) & tie ($\%$) & lose ($\%$) \\
  \hline
   w/o quality-tuning & 24.8 & 1.4 & 73.9 \\
                       100 & 60.3 & 1.5 & 38.2\\ 
                       1000 & 63.2 & 1.9 & 35.0\\ 
                       2000 & 67.0 & 2.6 & 30.4\\
 \end{tabular}
  \caption{\textbf{Visual Appeal by Fine-Tuning Dataset Size.} All the numbers are against SDXL as baseline. With  merely 100 high-quality images as fine-tuning data, our quality-tuned model can already outperform SDXL in visual appeal. Our model's visual appeal further improves as more images are used.
  }
  \vspace{-0.5cm}
\label{tab:dataset_size_visual} 
\end{table}

\section{Limitation}
\noindent \textbf{Limitation of Human Evaluation.} Relative to most published works, the prompt sets we evaluate on are reasonably large (1600 Parti prompts and our 2100 Open User Input prompts), under a multi-review annotation setting. Even then, the evaluation results may not fully reflect real-world usage of the models. In addition, human evaluation of text-to-image models, especially when it comes to aesthetics, is inherently subjective and noisy. As a result, evaluation on a different set of prompts or with different annotators and guidelines may lead to different results.
\vspace{0.5mm}

\noindent \textbf{Limitations of Small-Scale Fine-Tuning.} The role of quality-tuning is to restrict the output distribution to a high-quality domain. However, issues rooted from pre-training may still persist. For instance, the models may struggle to generate certain objects that were not sufficiently learned during pre-training.
\vspace{3mm}

\noindent \textbf{Limitations of Text-to-Image Models in General.} Like other text-to-image models, our models may generate biased, misleading, or offensive outputs. We've invested a significant amount in fairness and safety of our models - starting with balanced dataset construction, creating dedicated evaluation sets for these risk categories and investing multiple hours of redteaming.

\section{Conclusion}
In this paper, we demonstrated that manually selecting high quality images that are highly aesthetically-pleasing is one of the most important keys to improving the aesthetics of images generated by text-to-image generative models.  
We showed that with just a few hundred to thousand fine-tuning images, we were able to improve the visual appeal of generated images without compromising on the generality of visual concepts depicted. With this finding, we build \OURS, a LDM for high-quality image synthesis. On the commonly used PartiPrompts  and our Open User Input Prompts, we carefully evaluate our model against a publicly available state-of-the-art text-to-image model (SDXLv1.0~\cite{podell2023sdxl}) as well as our pre-trained LDM. We also show that quality-tuning not only improves LDMs, but also pixel diffusion and masked generative transformer models.
\section{Acknowledgement} 
This work would not have been possible without a large group of collaborators who helped with the underlying infrastructure, data, privacy, and the evaluation framework. We extend our gratitude to the following people for their contributions (alphabetical order): Eric Alamillo, Andrés Alvarado, Giri Anantharaman, Stuart Anderson, Snesha Arumugam, Chris Bray, Matt Butler, Anthony Chen, Lawrence Chen, Jessica Cheng, Lauren Cohen, Jort Gemmeke, Freddy Gottesman, Nader Hamekasi, Zecheng He, Jiabo Hu, Praveen Krishnan, Carolyn Krol, Tianhe Li, Mo Metanat, Vivek Pai, Guan Pang, Albert Pumarola, Ankit Ramchandani, Stephen Roylance, Kalyan Saladi, Artsiom Sanakoyeu, Dev Satpathy, Alex Schneidman, Edgar Schoenfeld, Shubho Sengupta, Hardik Shah, Shivani Shah, Yaser Sheikh, Karthik Sivakumar, Lauren Spencer, Fei Sun, Ali Thabet, Mor Tzur, Mike Wang, Mack Ward, Bichen Wu, Seiji Yamamoto, Licheng Yu, Hector Yuen, Luxin Zhang, Yinan Zhao, and Jessica Zhong. 

Finally, thank you Connor Hayes, Manohar Paluri and Ahmad Al-Dahle for your support and leadership.

{\small
\bibliographystyle{ieee_fullname}
\bibliography{egbib}

\begin{thebibliography}{10}\itemsep=-1pt

\bibitem{gpt4}
https://cdn.openai.com/papers/gpt-4.pdf.

\bibitem{chatgpt}
https://openai.com/blog/chatgpt/.

\bibitem{noiseoffset}
https://www.crosslabs.org/blog/diffusion-with-offset-noise/.

\bibitem{aghajanyan2022cm3}
Armen Aghajanyan, Bernie Huang, Candace Ross, Vladimir Karpukhin, Hu Xu, Naman Goyal, Dmytro Okhonko, Mandar Joshi, Gargi Ghosh, Mike Lewis, et~al.
\newblock Cm3: A causal masked multimodal model of the internet.
\newblock {\em arXiv preprint arXiv:2201.07520}, 2022.

\bibitem{balaji2022ediffi}
Yogesh Balaji, Seungjun Nah, Xun Huang, Arash Vahdat, Jiaming Song, Karsten Kreis, Miika Aittala, Timo Aila, Samuli Laine, Bryan Catanzaro, et~al.
\newblock ediffi: Text-to-image diffusion models with an ensemble of expert denoisers.
\newblock {\em arXiv preprint arXiv:2211.01324}, 2022.

\bibitem{bender2021dangers}
Emily~M Bender, Timnit Gebru, Angelina McMillan-Major, and Shmargaret Shmitchell.
\newblock On the dangers of stochastic parrots: Can language models be too big?
\newblock In {\em Proceedings of the 2021 ACM conference on fairness, accountability, and transparency}, pages 610--623, 2021.

\bibitem{bommasani2021opportunities}
Rishi Bommasani, Drew~A Hudson, Ehsan Adeli, Russ Altman, Simran Arora, Sydney von Arx, Michael~S Bernstein, Jeannette Bohg, Antoine Bosselut, Emma Brunskill, et~al.
\newblock On the opportunities and risks of foundation models.
\newblock {\em arXiv preprint arXiv:2108.07258}, 2021.

\bibitem{brooks2023instructpix2pix}
Tim Brooks, Aleksander Holynski, and Alexei~A Efros.
\newblock Instructpix2pix: Learning to follow image editing instructions.
\newblock In {\em Proceedings of the IEEE/CVF Conference on Computer Vision and Pattern Recognition}, pages 18392--18402, 2023.

\bibitem{chang2023muse}
Huiwen Chang, Han Zhang, Jarred Barber, AJ Maschinot, Jose Lezama, Lu Jiang, Ming-Hsuan Yang, Kevin Murphy, William~T Freeman, Michael Rubinstein, et~al.
\newblock Muse: Text-to-image generation via masked generative transformers.
\newblock {\em arXiv preprint arXiv:2301.00704}, 2023.

\bibitem{ding2021cogview}
Ming Ding, Zhuoyi Yang, Wenyi Hong, Wendi Zheng, Chang Zhou, Da Yin, Junyang Lin, Xu Zou, Zhou Shao, Hongxia Yang, et~al.
\newblock Cogview: Mastering text-to-image generation via transformers.
\newblock {\em Advances in Neural Information Processing Systems}, 34:19822--19835, 2021.

\bibitem{gafni2022make}
Oran Gafni, Adam Polyak, Oron Ashual, Shelly Sheynin, Devi Parikh, and Yaniv Taigman.
\newblock Make-a-scene: Scene-based text-to-image generation with human priors.
\newblock In {\em European Conference on Computer Vision}, pages 89--106. Springer, 2022.

\bibitem{gal2022image}
Rinon Gal, Yuval Alaluf, Yuval Atzmon, Or Patashnik, Amit~H Bermano, Gal Chechik, and Daniel Cohen-Or.
\newblock An image is worth one word: Personalizing text-to-image generation using textual inversion.
\newblock {\em arXiv preprint arXiv:2208.01618}, 2022.

\bibitem{gehman2020realtoxicityprompts}
Samuel Gehman, Suchin Gururangan, Maarten Sap, Yejin Choi, and Noah~A Smith.
\newblock Realtoxicityprompts: Evaluating neural toxic degeneration in language models.
\newblock {\em arXiv preprint arXiv:2009.11462}, 2020.

\bibitem{ho2020denoising}
Jonathan Ho, Ajay Jain, and Pieter Abbeel.
\newblock Denoising diffusion probabilistic models.
\newblock {\em Advances in neural information processing systems}, 33:6840--6851, 2020.

\bibitem{hu2021lora}
Edward~J Hu, Yelong Shen, Phillip Wallis, Zeyuan Allen-Zhu, Yuanzhi Li, Shean Wang, Lu Wang, and Weizhu Chen.
\newblock Lora: Low-rank adaptation of large language models.
\newblock {\em arXiv preprint arXiv:2106.09685}, 2021.

\bibitem{huang2023noise2music}
Qingqing Huang, Daniel~S Park, Tao Wang, Timo~I Denk, Andy Ly, Nanxin Chen, Zhengdong Zhang, Zhishuai Zhang, Jiahui Yu, Christian Frank, et~al.
\newblock Noise2music: Text-conditioned music generation with diffusion models.
\newblock {\em arXiv preprint arXiv:2302.03917}, 2023.

\bibitem{kang2023scaling}
Minguk Kang, Jun-Yan Zhu, Richard Zhang, Jaesik Park, Eli Shechtman, Sylvain Paris, and Taesung Park.
\newblock Scaling up gans for text-to-image synthesis.
\newblock In {\em Proceedings of the IEEE/CVF Conference on Computer Vision and Pattern Recognition}, pages 10124--10134, 2023.

\bibitem{kirstain2023pick}
Yuval Kirstain, Adam Polyak, Uriel Singer, Shahbuland Matiana, Joe Penna, and Omer Levy.
\newblock Pick-a-pic: An open dataset of user preferences for text-to-image generation.
\newblock {\em arXiv preprint arXiv:2305.01569}, 2023.

\bibitem{lin2023magic3d}
Chen-Hsuan Lin, Jun Gao, Luming Tang, Towaki Takikawa, Xiaohui Zeng, Xun Huang, Karsten Kreis, Sanja Fidler, Ming-Yu Liu, and Tsung-Yi Lin.
\newblock Magic3d: High-resolution text-to-3d content creation.
\newblock In {\em Proceedings of the IEEE/CVF Conference on Computer Vision and Pattern Recognition}, pages 300--309, 2023.

\bibitem{ouyang2022training}
Long Ouyang, Jeffrey Wu, Xu Jiang, Diogo Almeida, Carroll Wainwright, Pamela Mishkin, Chong Zhang, Sandhini Agarwal, Katarina Slama, Alex Ray, et~al.
\newblock Training language models to follow instructions with human feedback.
\newblock {\em Advances in Neural Information Processing Systems}, 35:27730--27744, 2022.

\bibitem{podell2023sdxl}
Dustin Podell, Zion English, Kyle Lacey, Andreas Blattmann, Tim Dockhorn, Jonas M{\"u}ller, Joe Penna, and Robin Rombach.
\newblock Sdxl: Improving latent diffusion models for high-resolution image synthesis.
\newblock {\em arXiv preprint arXiv:2307.01952}, 2023.

\bibitem{poole2022dreamfusion}
Ben Poole, Ajay Jain, Jonathan~T Barron, and Ben Mildenhall.
\newblock Dreamfusion: Text-to-3d using 2d diffusion.
\newblock {\em arXiv preprint arXiv:2209.14988}, 2022.

\bibitem{radford2021learning}
Alec Radford, Jong~Wook Kim, Chris Hallacy, Aditya Ramesh, Gabriel Goh, Sandhini Agarwal, Girish Sastry, Amanda Askell, Pamela Mishkin, Jack Clark, et~al.
\newblock Learning transferable visual models from natural language supervision.
\newblock In {\em International conference on machine learning}, pages 8748--8763. PMLR, 2021.

\bibitem{raffel2020exploring}
Colin Raffel, Noam Shazeer, Adam Roberts, Katherine Lee, Sharan Narang, Michael Matena, Yanqi Zhou, Wei Li, and Peter~J Liu.
\newblock Exploring the limits of transfer learning with a unified text-to-text transformer.
\newblock {\em The Journal of Machine Learning Research}, 21(1):5485--5551, 2020.

\bibitem{ramesh2022hierarchical}
Aditya Ramesh, Prafulla Dhariwal, Alex Nichol, Casey Chu, and Mark Chen.
\newblock Hierarchical text-conditional image generation with clip latents.
\newblock {\em arXiv preprint arXiv:2204.06125}, 2022.

\bibitem{ramesh2021zero}
Aditya Ramesh, Mikhail Pavlov, Gabriel Goh, Scott Gray, Chelsea Voss, Alec Radford, Mark Chen, and Ilya Sutskever.
\newblock Zero-shot text-to-image generation.
\newblock In {\em International Conference on Machine Learning}, pages 8821--8831. PMLR, 2021.

\bibitem{rombach2022high}
Robin Rombach, Andreas Blattmann, Dominik Lorenz, Patrick Esser, and Bj{\"o}rn Ommer.
\newblock High-resolution image synthesis with latent diffusion models.
\newblock In {\em Proceedings of the IEEE/CVF conference on computer vision and pattern recognition}, pages 10684--10695, 2022.

\bibitem{ruiz2022dreambooth}
Nataniel Ruiz, Yuanzhen Li, Varun Jampani, Yael Pritch, Michael Rubinstein, and Kfir Aberman.
\newblock Dreambooth: Fine tuning text-to-image diffusion models for subject-driven generation.
\newblock 2022.

\bibitem{russakovsky2015imagenet}
Olga Russakovsky, Jia Deng, Hao Su, Jonathan Krause, Sanjeev Satheesh, Sean Ma, Zhiheng Huang, Andrej Karpathy, Aditya Khosla, Michael Bernstein, et~al.
\newblock Imagenet large scale visual recognition challenge.
\newblock {\em International journal of computer vision}, 115:211--252, 2015.

\bibitem{saharia2022photorealistic}
Chitwan Saharia, William Chan, Saurabh Saxena, Lala Li, Jay Whang, Emily~L Denton, Kamyar Ghasemipour, Raphael Gontijo~Lopes, Burcu Karagol~Ayan, Tim Salimans, et~al.
\newblock Photorealistic text-to-image diffusion models with deep language understanding.
\newblock {\em Advances in Neural Information Processing Systems}, 35:36479--36494, 2022.

\bibitem{sauer2023stylegan}
Axel Sauer, Tero Karras, Samuli Laine, Andreas Geiger, and Timo Aila.
\newblock Stylegan-t: Unlocking the power of gans for fast large-scale text-to-image synthesis.
\newblock {\em arXiv preprint arXiv:2301.09515}, 2023.

\bibitem{singer2022make}
Uriel Singer, Adam Polyak, Thomas Hayes, Xi Yin, Jie An, Songyang Zhang, Qiyuan Hu, Harry Yang, Oron Ashual, Oran Gafni, et~al.
\newblock Make-a-video: Text-to-video generation without text-video data.
\newblock {\em arXiv preprint arXiv:2209.14792}, 2022.

\bibitem{touvron2023llama}
Hugo Touvron, Thibaut Lavril, Gautier Izacard, Xavier Martinet, Marie-Anne Lachaux, Timoth{\'e}e Lacroix, Baptiste Rozi{\`e}re, Naman Goyal, Eric Hambro, Faisal Azhar, et~al.
\newblock Llama: Open and efficient foundation language models.
\newblock {\em arXiv preprint arXiv:2302.13971}, 2023.

\bibitem{wang2023prolificdreamer}
Zhengyi Wang, Cheng Lu, Yikai Wang, Fan Bao, Chongxuan Li, Hang Su, and Jun Zhu.
\newblock Prolificdreamer: High-fidelity and diverse text-to-3d generation with variational score distillation.
\newblock {\em arXiv preprint arXiv:2305.16213}, 2023.

\bibitem{weidinger2021ethical}
Laura Weidinger, John Mellor, Maribeth Rauh, Conor Griffin, Jonathan Uesato, Po-Sen Huang, Myra Cheng, Mia Glaese, Borja Balle, Atoosa Kasirzadeh, et~al.
\newblock Ethical and social risks of harm from language models.
\newblock {\em arXiv preprint arXiv:2112.04359}, 2021.

\bibitem{yalniz2019billion}
I~Zeki Yalniz, Herv{\'e} J{\'e}gou, Kan Chen, Manohar Paluri, and Dhruv Mahajan.
\newblock Billion-scale semi-supervised learning for image classification.
\newblock {\em arXiv preprint arXiv:1905.00546}, 2019.

\bibitem{yu2022scaling}
Jiahui Yu, Yuanzhong Xu, Jing~Yu Koh, Thang Luong, Gunjan Baid, Zirui Wang, Vijay Vasudevan, Alexander Ku, Yinfei Yang, Burcu~Karagol Ayan, et~al.
\newblock Scaling autoregressive models for content-rich text-to-image generation.
\newblock {\em arXiv preprint arXiv:2206.10789}, 2(3):5, 2022.

\bibitem{yu2023scaling}
Lili Yu, Bowen Shi, Ramakanth Pasunuru, Benjamin Muller, Olga Golovneva, Tianlu Wang, Arun Babu, Binh Tang, Brian Karrer, Shelly Sheynin, et~al.
\newblock Scaling autoregressive multi-modal models: Pretraining and instruction tuning.
\newblock {\em arXiv preprint arXiv:2309.02591}, 2023.

\bibitem{zhang2023adding}
Lvmin Zhang and Maneesh Agrawala.
\newblock Adding conditional control to text-to-image diffusion models.
\newblock {\em arXiv preprint arXiv:2302.05543}, 2023.

\end{thebibliography}
}

\end{document}